\documentclass[fleqn,10pt]{wlscirep}
\usepackage[utf8]{inputenc}
\usepackage[T1]{fontenc}
\usepackage{titlesec}
\usepackage{graphicx}
\usepackage{array}
\usepackage{caption}
\usepackage{subcaption}
\usepackage{etoolbox,refcount}
\usepackage{multicol}
\usepackage{fontawesome}
\usepackage{hyperref}
\usepackage{nameref}
\usepackage{verbatim}
\usepackage{calculus}
\usepackage{amsmath}
\usepackage{nameref}
\usepackage{amsmath}

\title{Quantitative Evaluation of the Severity of Posttraumatic Stress Disorder through Transfer Learning from Specific Phobia Data}
\author[1,*]{Nicolas RICKA}
\author[1]{Gauthier PELLEGRIN}
\author[1]{Denis A. FOMPEYRINE}
\author[2]{Thomas ROHALY}
\author[3]{Leah ENDERS}
\author[3]{Heather ROY}

\affil[1]{MyndBlue, F-75008 Paris, France}
\affil[2]{DCS Corporation, Alexandria, Virginia, United States of America}
\affil[3]{Human in Complex Systems Division, DEVCOM Army Research Laboratory, Aberdeen Proving Ground, Maryland, United States of America}
\affil[*]{Corresponding author, \texttt{nicolas@myndblue.ai}}
\date{December 2025}

\begin{abstract}  
Posttraumatic stress disorder (PTSD) is a prevalent and debilitating mental health condition with significant personal and societal impacts. Current clinical assessments of PTSD often rely on subjective evaluations, which can be time-consuming, costly, and prone to human bias.
This study proposes a machine learning (ML) approach based on multivariate kernel density estimation (MKDE) technique for the objective evaluation of PTSD severity. We collected heart rate (HR) and galvanic skin response (GSR) signals as well as PTSD Checklist - Military Version (PCL-M) labels from 21 participants during an immersive simulation. A fear-response model was trained on a public arachnophobia dataset, and predictive features of PTSD were extracted from the fear-response curves estimated on the military dataset.
The model achieved an accuracy of 86\% in classifying PTSD status, effectively distinguishing participants with and without PTSD (PCL-M threshold of 36). The average mean absolute error (MAE) of the models is 5.6, and it estimated a clinical PTSD severity scale with a mean absolute percentage error of 17\%.
Our algorithm demonstrates promising potential for enhancing estimation of PTSD severity and followup by offering an objective and low-effort evaluation approach using physiology. These findings suggest clinical utility in both screening and follow-up settings.
\vspace{.2cm}
\end{abstract}

\begin{document}

\maketitle


\section*{Introduction}
Posttraumatic stress disorder (PTSD) is a mental health condition that may develop in individuals exposed to traumatic events involving death, serious injury, or sexual assault \cite{dsm5}. PTSD is a global public health concern, with a cross-national lifetime prevalence of 3.9\% and 5.6\% among the trauma exposed populations \cite{Koe17}, and the prevalence of PTSD vary widely across regions, populations, and trauma types \cite{Ben16}. Large-scale epidemiological studies have shown that trauma exposure is common worldwide and that PTSD prevalence can be particularly high in regions affected by conflict, political instability, or disasters, such as Sub-Saharan Africa, South Asia, and the Middle East \cite{Koe17, Cha19, deJon01}. Even among countries spared by these problems, with comparatively lower prevalence and mainly explained by other factors such as child abuse or sexual assault, the picture is quite diverse in terms of prevalence, affected populations, and traumatic experiences causing PTSD \cite{Rze23}. This broader epidemiological perspective underscores the need for objective and scalable tools to assist PTSD screening across a range of populations and contexts. While our study focuses on military-related PTSD, where lifetime prevalence has been estimated as high as 7\% for male U.S. veterans and 12\% for female U.S veterans \cite{VA_PTSD}, the approach we explore may offer insights for future work in other trauma-exposed populations.
PTSD can be difficult to assess because of its complex clinical manifestations, and the methods used to assess it currently rely mostly on subjective assessments \cite{roberts2022, bovin2015, resick2023}. 
Objective tests, which are mainly biological and based on blood or saliva samples, are available, but they are time-consuming and costly, making these methods unsuitable for being scaled to test a larger population \cite{schein2021, greene2016, Wu23}. 
The relationship between physiology and PTSD has been known for several decades, and our understanding of this complex connection has been refined over the years. Early work by Shalev and Rogel-Fuchs \cite{Sha93} and Krystal \textit{et al.} \cite{Kry89} laid the foundation for understanding the neurobiological aspects of PTSD, with a focus on the role of central noradrenergic systems. This area was further explored by Kuch \textit{et al.}\cite{Kuch96}, who reported that psychophysiological responses, such as heart rate and blood pressure, could distinguish between individuals with PTSD and those without PTSD. These psychophysiological approaches to PTSD diagnosis have provided valuable objective indicators of stress and trauma response \cite{Pol07, Orr00}. However, without machine learning, these methods typically require structured laboratory tasks and manual signal interpretation, limiting their scalability and real-world applicability. Recent advances in machine learning as well as in wearable monitoring devices offer a promising alternative by enabling automated, real-time analysis of minimally processed physiological signals collected in naturalistic or ambulatory contexts. These tools also enable to transfer knowledge from one dataset to the other. For instance fine tuning model can enhance generalizability across populations and trauma types, ultimately increasing the diagnostic accuracy. Alternatively, transferring the learned relationships between physiology and diagnostic between closely related diseases is not possible without machine learning.
Earlier publication by the authors on the same database \cite{GERM} focused on the discrimination between PTSD and no-PTSD subjects by analyzing the habituation in response to specific events in the simulation. In contrast, the current analysis addresses a different research question, applying transfer learning from a specific phobia to investigate PTSD symptom intensity evaluation, informing on the disease's severity, without using specific events in the database. Thus, the work presented here addresses a different scientific questions, with different methods, and there is no overlap between the two studies. Shalev \cite{Sha16} provided a comprehensive overview of the field, highlighting the contributions of trauma characteristics and genetic, biological, and psychosocial risk factors to the development of PTSD. 

The use of wearables for the medical estimation of physiological characteristics has been steadily increasing since the last decade, with a focus on wrist-worn devices, and is still being improved today \cite{Miy24}. These devices can be equipped with various sensors, such as accelerometers, photoplethysmograph (PPG) or galvanic skin response (GSR) sensors, which can accurately estimate physical activity, cardiorespiratory features and parasympathetic activity, as described in \cite{Hen18}.

Linking PTSD with physiological data related to disorders with similar fear  or anxiety based responses, such as specific phobia, can provide additional information. Specific phobia is marked by fear or anxiety related to a specific object or situation. Previous studies have suggested a significant degree of similarity between phobic reactions and PTSD, with both conditions involving fear and avoidance behaviors. Functional neuroimaging studies have revealed commonalities in brain mechanisms, particularly increased activity in the amygdala and insula, between individuals with normal fear responses and those with anxiety disorders, including PTSD and some specific phobia \cite{Etkin07}. Similarly, reduced specificity in autobiographical memory retrieval is associated with the development of both disorders \cite{Kle08}.

However, PTSD and specific phobia show some important differences. For example, intrusive thoughts and memories occur more frequently in patients with PTSD than in phobic individuals \cite{Pfa13}. Moreover, PTSD patients exhibit additional abnormalities in the regions of emotion regulation, which may explain symptoms that go beyond exaggerated fear responses \cite{Kle08}. 

Previous studies have shown the efficacy of machine learning (ML) in personalized healthcare \cite{zhe13}. Detection of PTSD is possible using sleeping patterns \cite{wan24}, which are available using wearable data; however, none of these results were obtained in short-term experiments, and the methods developed are not yet suitable for evaluating the severity of PTSD \cite{tah17}. Our work fills this gap in the literature and strengthens our understanding of the relationship between specific phobias and PTSD.

The purpose of this study is twofold. First, this work serves as a test case for estimating PTSD severity, measured using the PTSD Checklist - Military Version (PCL-M), solely from physiological data collected passively during a short immersive simulation. Second, we leveraged the  known similarities between PTSD and specific phobia disorders by transferring the estimate of fear responses related to a specific phobia to estimate the severity of PTSD.

\begin{figure*}[!ht]
    \centering
    
    \includegraphics[scale=0.3]{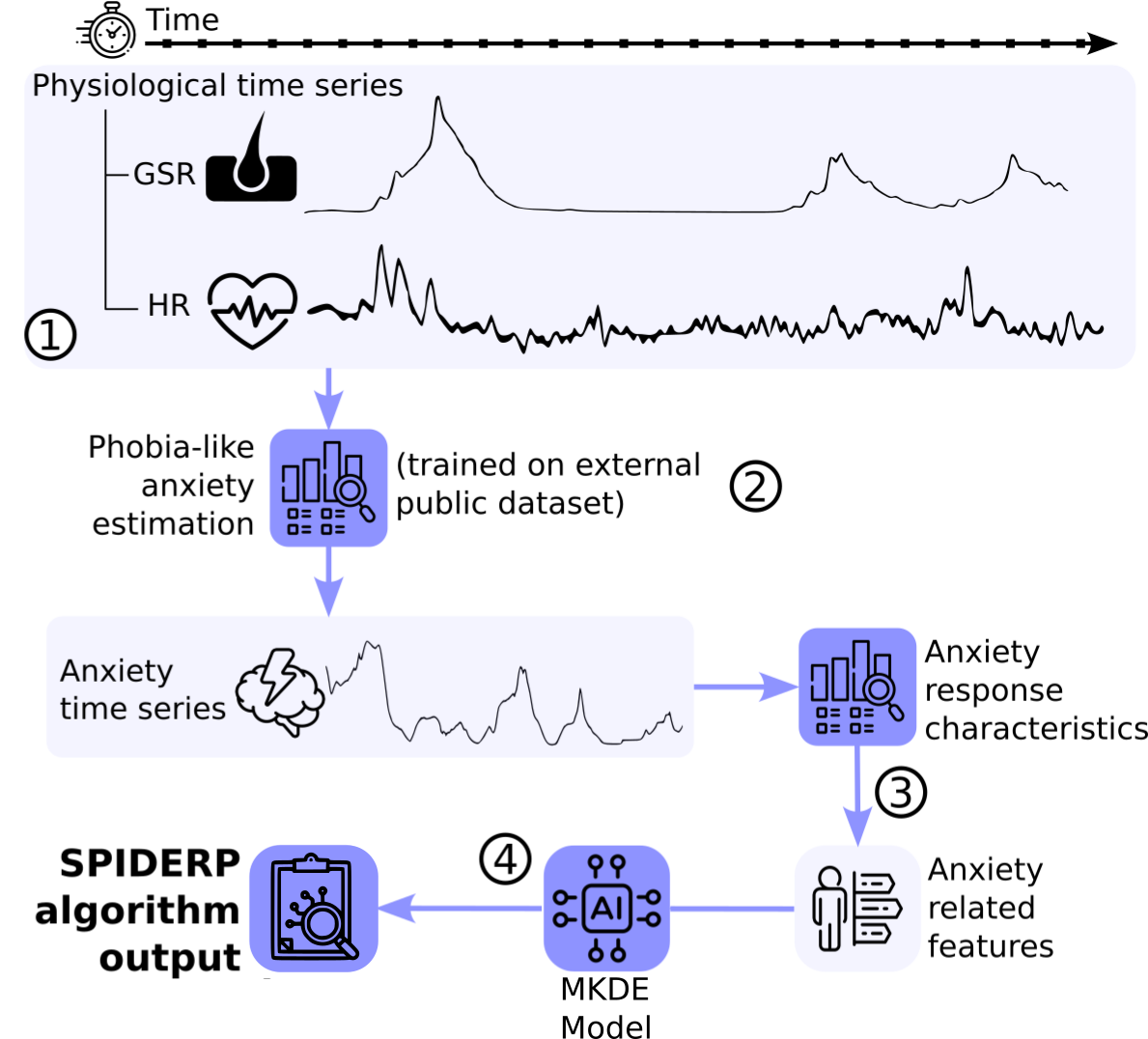}
    \caption{Overview of the proposed algorithm, SPIDERP. The model reads sensor measurement time series, and outputs a probability curve for the different possible outcomes. The data flow is divided into 4 steps: physiology extraction, estimation of the fear response, computation of static fear response features, and the multivariate kernel density estimator (MKDE) providing the final prediction.}
    \label{fig:SPIDERPOverview}
\end{figure*}

Our Stress PTSD Indicator Derived from Events Reaction in Physiology (SPIDERP) algorithm provides a novel approach for detecting PTSD. This algorithm is divided into 2 main components. The first one, the fear response (FR) model, assigns a score between 0 and 1, representing whether samples composed of 20 s windows of physiological data indicate that a phobic subject is currently experiencing a fear response, i.e. a manifestation of their phobia. This is similar to the scored called "anxiety scores" in \cite{ArachnoAnxietyDetection20}, since PTSD overlaps significantly with anxiety disorders \cite{Mae17}. This estimate of fear response is used to detect a key components of PTSD: fear dysregulation \cite{Kre22} involving extreme and remarkable arousal \cite{Pfa13}. This model outputs a FR curve for each physiological record. The second component of SPIDERP uses fear response curves built from the previous model to derive indicators of PTSD severity. An overview of the strategy presented in this paper is depicted in Figure \ref{fig:SPIDERPOverview}.

Combining these components yield SPIDERP, a model that outputs a probability distribution over all possible PCL-M values from physiological data. This output can be reduced to a single PCL-M prediction by selecting the PCL-M value with the maximum probability according to the model. Finally, PCL-M values can be further reduced to a binary output representing PTSD status using a threshold according to PCL-M \cite{pcl_thresh}.

\section*{Materials and methods}

\subsection*{Target dataset: military PTSD}
The study protocol and all methods were carried out in accordance with the accredited Institutional Review Board (IRB) at the U.S. Army Combat Capabilities Development Command Army Research Laboratory (ARL), who approved the study, and conducted in compliance with the ARL Human Research Protection Program (HRPP) and the Declaration of Helsinki. The protocol was approved by the ARL HRPP (ARL 21–104). Data were collected from 41 active-duty U.S. service members and veterans recruited from Joint Base San Antonio, Texas. The recruitment period spanned from initial approval on September 15, 2021, until study closure on November 1, 2023. Written informed consent was obtained from all participants, and all participants reviewed and signed an IRB approved consent form prior to participation. 
The participants were tasked with completing a series of visual search tasks within an immersive desktop environment with military stimuli (e.g., military vehicles) and events (e.g., gunfire, flashbangs) aimed at inducing a high stress state. Participants engaged in the visual search task, searching for an assigned target object (e.g., green SUV) for approximately 30 minutes in total. Electrocardiogram (ECG), and GSR data (both recorded at 2048 Hz through the Biosemi Active two System) were collected throughout the task. See \cite{STR24} for a full description of the task protocol and a complete list of events. 
Participants filled out the self-reported PTSD Checklist for Military version (PLC-M), resulting in a label between 17 and 85 representing PTSD symptoms severity. Additionally, participants were assigned a binary label PTSD or no PTSD based on a threshold set to 36 \cite{pcl_thresh} applied to the PCL-M score \cite{Weathers1993}. The demographic and clinical characteristics of the participants are detailed in Table \ref{tab:demographics}. Among the 41 participants included, 4 did not complete the PCL-M questionnaire, 12 had one or more incorrectly measured physiological signals (either due to a technical problem during the experiment or corrupted data saving), and 4 had no events recorded during the simulation. The 21 remaining participants were considered in our results and analysis. The cohort was 90.5\% male, the average age was 51.4 $\pm$ 13.9 years, and the median age was 54 (range: 26–69) years. The average PCL-M score was 32.5 $\pm$ 13.1, with a median score of 29 (range: 17–54). Among this population, 61.9\% were veterans and 57.14\% had previous combat experience (see Table \ref{tab:demographics} for full baseline demographics of our cohort).

\begin{table*}[t!]
\begin{center}
\begin{tabular}{|l|l|}
\hline
\textbf{Variables} & \\ \hline\hline

\rowcolor{lightgray}
Mean ($\pm$SD) age, years & 51.4 $\pm$ 13.9 \\
\rowcolor{lightgray}
Median (min-max) age, years & 54 (26-69) \\\hline

Mean ($\pm$SD) PCLM score & 32.5 $\pm$ 13.1 \\
Median (min-max) PCLM score & 29 (17-54) \\\hline

\rowcolor{lightgray}
Combat experience, yes:no, n (\%) & 12 (57.1\%) : 9 (42.9\%) \\\hline

Mild Traumatic Brain Injury, yes:no (\%) & 3 (14.3\%) : 18 (85.7\%) \\\hline

\rowcolor{lightgray}
Mean ($\pm$SD) deployments, number & 2.1 $\pm$ 2.3 \\
\rowcolor{lightgray}
Median (min-max) deployments, number & 2 (0-10) \\\hline

Sex, M:F, n (\%) & 19 (90.5\%) : 2 (9.5\%)  \\\hline

\rowcolor{lightgray}
Mean ($\pm$SD) time in service, years & 18.2 $\pm$ 9.7 \\
\rowcolor{lightgray}
Median (min-max) time in service, years & 20 (1-36) \\\hline

Is Veteran, yes:no, n (\%) & 13 (61.9\%) : 8 (38.1\%) \\\hline

\rowcolor{lightgray}
Is officer, yes:no:missing, n (\%) &  2 (9.5\%) : 16 (76.2\%) : 3 (14.3\%) \\\hline

Has PTSD, yes:no, n (\%) & 11 (52.4\%) : 10 (47.6\%) \\\hline
\end{tabular}
\caption{Baseline demographics and clinical characteristics.}
\label{tab:demographics}
\end{center}
\end{table*}

\subsection*{Source dataset: arachnophobia}

The dataset used to model phobic reactions to spiders is from the work of Ihmig \textit{et al.} \cite{ArachnoDataset20}, and the reader can refer to the original publication for an extensive description of the database and data collection procedure. In this work, we refer to this dataset as the arachnophobia dataset. 
For a detailed description of the study protocol, see \cite{ArachnoAnxietyDetection20}.

To enable comparison with the PTSD database, we retained only the ECG and GSR channels, which are present in both datasets. 
The dataset contains records from 57 subjects, and each record is approximately 35 minutes long. Data were segmented into 20-second windows, and labeled with $FR = 1$ if the sample corresponded to a fear response and $FR = 0$ otherwise.

\subsection*{Physiology preprocessing steps and vectorization}

The same processing steps were applied to both the PTSD and Arachnophobia datasets. First, we extracted 2 physiological signals from the databases: heart rate (HR) from ECG data using standard processing methods, and phasic GSR, which correspond to rapid, transient changes in skin conductance that occur in response to specific stimuli or events from the raw GSR measurements.
Each signal was subsequently subjectwise normalized, and divided into 20 s sliding windows, with a stride of 1 s. Finally, for each window and each physiological signal, we computed 4 features: normalized mean, normalized std, normalized first difference, and normalized second difference (see Table \ref{tab:feature_construction} in the Supplementary material for more details). This yields a 12-dimensional dataset.

\subsection*{Fear response model training}

We trained a multilayer perceptron (MLP) to predict FR labels (see Table \ref{tab:hyperparameters} for the chosen hyperparameters).

The arachnophobia dataset was split into training and testing datasets using K-fold stratified by subject to account for inter-subject heterogeneity. This results in a set of K trained FR models.

At inference time, we average the output scores obtained by the K models. We emphasize that the K models are aggregated by averaging the floating point numbers and not by a majority vote, as is often used in binary predictions via ensemble models.

\subsection*{Static fear response features}

Once the ML model for FR prediction has been trained, the next step is to consider the PTSD dataset. Given the physiological time series associated with a subject in this dataset, using the procedure described above, we obtain a fear response curve $A_t$. For different subjects, these fear response curves are of different lengths depending on the duration of the experiment. 

Considering the small number of subjects, to avoid the curse of dimensionality while predicting the PCL-M value, we computed a set of features from these time series:
\begin{itemize}
    \item \textit{fear response slope}: slope obtained by fitting a linear regression on the fear response curve,
    \item \textit{initial fear response}: mean fear response during the first 30 seconds of the experiment,
    \item \textit{sex}: boolean variable, with values of 0 (male) and 1 (female).
\end{itemize}
While the physiological data producing the features used in our model (e.g., heart rate and electrodermal activity) are grounded in traditional psychophysiological markers of PTSD \cite{Pol07}, our approach diverges by applying data-driven machine learning to these signals. This allows for automated detection of PTSD without the need for manual thresholding of task-evoked responses, supporting broader scalability and deployment.

\subsection*{PTSD model description}

We used a multivariate kernel density estimation (MKDE) model to predict PCL-M. This density estimation model, where the density functions are estimated over each variable individually and then aggregated, learns the probability distribution on the possible PCL-M values: $P(y)$ for $y \in [17, 85].$ 

Given a sample with feature vector $X$ and label $y$, the MKDE-based prediction function is defined as follows:

$$ P(y | X = x) =\frac{1}{M} \sum_{m=0}^{M-1} K_y(y, y_m) * K_X(x, x_m),$$
where $K_x$, $K_y$ are the kernel functions (gaussian kernel for continuous variables and bernouilli kernels for binary variables).
Gaussian kernels depend on a single parameter $\sigma$, the bandwidth of the kernel, which is chosen to maximize likelihood over the train set.
The details of the construction of MKDE are provided in the Supplementary material.
Since the dataset contains only 21 subjects, separation into training and test sets is performed via a leave-one-out (LOO) validation scheme.

\section*{Results}

\subsection*{Fear response model}
\begin{figure*}[!ht]
    \centering
    \includegraphics[width=0.6\linewidth]{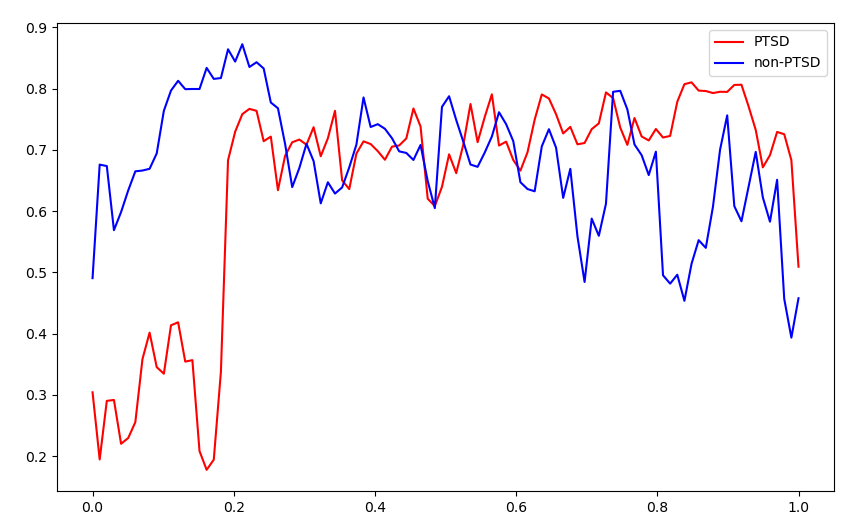}
    \caption{The median fear reaction curves for the PTSD (red) and non-PTSD (blue) subjects. We observe that the fear reactions of the non-PTSD patients decrease over time, whereas the PTSD patients have low initial fear reactions that drastically increase during the experiment. The time is in arbitrary units between 0 and 1, and the fear reaction is the output of the FR model, which is between 0 and 1.}
    \label{fig:anxiety_curves}
\end{figure*}

In the stratified K-fold validation scheme, our FR model had an accuracy of 70\% in distinguishing between fear response and rest samples. The model outputs on the featured physiological data from the PTSD dataset were then averaged to obtain the fear response curves, shown in Fig. \ref{fig:anxiety_curves}.

\subsection*{PTSD model}

The PTSD prediction model SPIDERP can be evaluated on two tasks: PTSD severity estimation and PTSD detection by reducing the PCL-M value to its corresponding binary label.

\begin{figure*}[ht]
    \centering
    \includegraphics[width=0.65\linewidth]{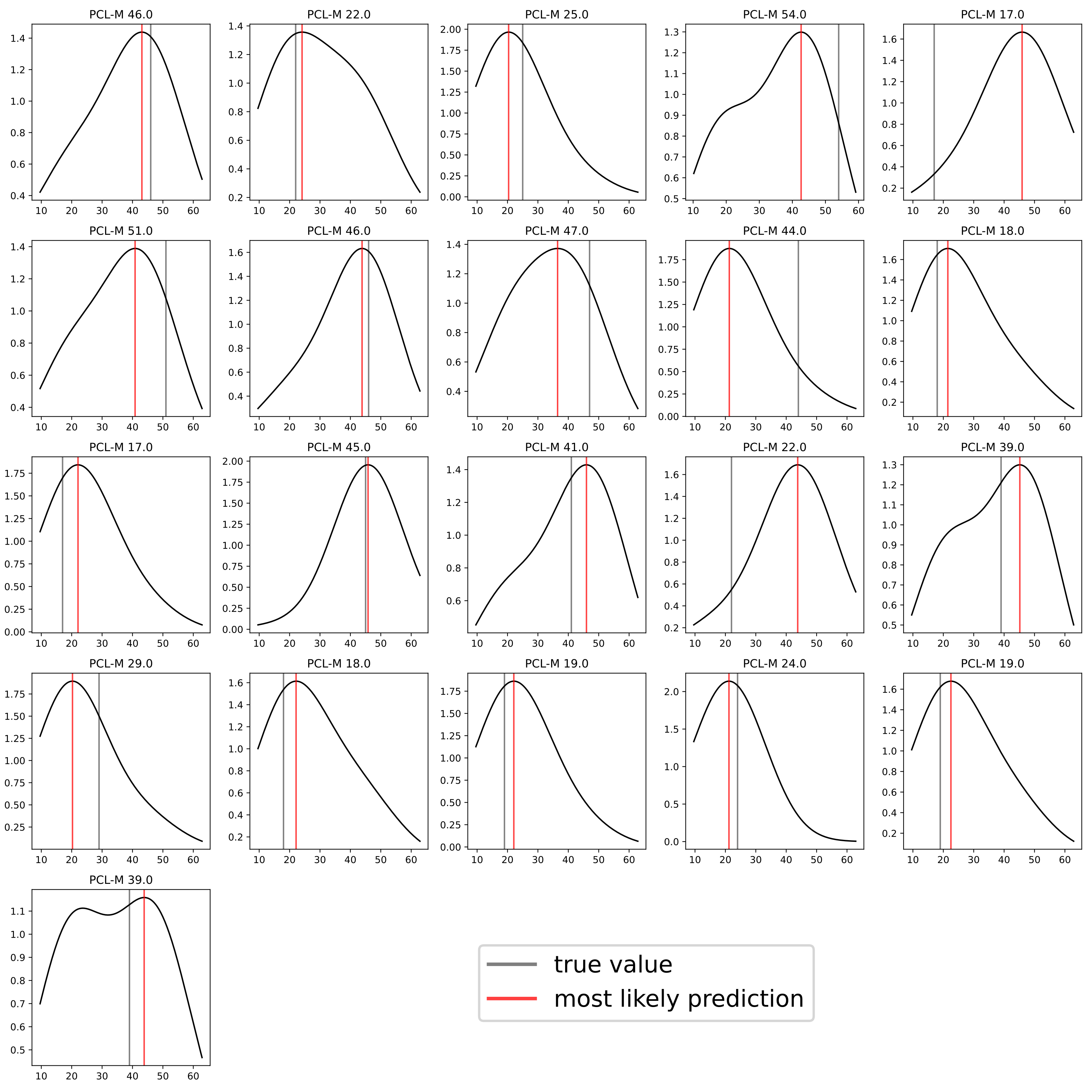}
    \caption{Raw outputs of the PTSD models for the individual subjects. The black curve represents the probability distribution, the gray vertical line represents the ground-truth PCL-M, and the red vertical line represents the most likely prediction according to the model.}
    \label{fig:predictions-proba}
\end{figure*}

As depicted in Fig. \ref{fig:predictions-proba}, every subject except the last one has a unique local maximum. The last subject has 2 local maxima, one at 22 and the other at 44.

The PTSD model have a mean absolute error (MAE) of 5.6 for determining the PCL-M value of subjects.

The mean absolute percentage error (MAPE) is 17\%. We compare the PTSD model to 2 different baselines, a naïve constant baseline predicting the mode of the distribution of samples, and a sex-specific baseline predicting the mode of the distribution of labels across participants of the same sex as the subject under consideration, see Table \ref{tab:results}.

\begin{table*}[t]
\begin{center}
\begin{tabular}{|l|l|l|l|}
\hline
\textbf{Metric} & \textbf{SPIDERP} &  \textbf{Constant baseline} & \textbf{Sex-specific baseline}\\ \hline\hline
\rowcolor{lightgray}
MAE & \textbf{5.6} & 11.2 & 10.7 \\
MAPE & \textbf{17\%} & 44.3\% & 39.4\% \\
\rowcolor{lightgray}
Binary accuracy & \textbf{86\%} & 52\% & 61\% \\  \hline
\end{tabular}
\caption{Comparison between our model and the chosen baselines: the mean-value baseline that predicts a PCL-M of 32 for all patients, and the sex-specific baseline that predicts 45.5 for females and 31.1 for males. The best results for each metric are in bold.}
\label{tab:results}
\end{center}
\end{table*}

\begin{figure*}[ht]
    \centering
    \includegraphics[width=0.4\linewidth]{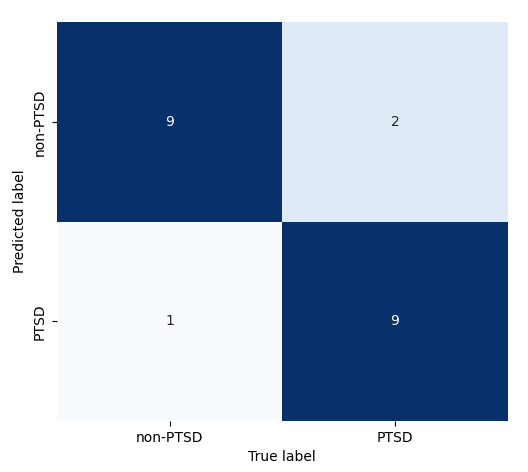}
    \caption{Confusion matrix for the 2-class classification problem. The model reaches an accuracy of 86\% in distinguishing between PTSD and non-PTSD subjects.} 
    \label{fig:confusion_matrix}
\end{figure*}

Reducing to binary label PTSD/non-PTSD, the PTSD model reached an accuracy of 86\%, with 3 classification errors. The corresponding confusion matrix is depicted in Fig. \ref{fig:confusion_matrix}.

\section*{Discussion}

Higher accuracies to distinguish between fear response and rest samples than ours (70\%) were reported in the literature \cite{Arachnophobia20}. This discrepancy can be explained by the necessities of our transfer learning approach.
\begin{enumerate}
    \item We restricted to common signals between the target and source databases, discarding signals used in other anxiety studies \cite{BRImportant21} and emotion analysis \cite{social22, Pic01} such as breathing rate.
    \item We used a simple subject-independent binary label from the arachnophobia dataset, in contrast to \cite{Arachnophobia20} where the authors considered subject specific label.
    \item Our preprocessing includes subject-wise normalization, essential for the transfer learning task considering the difference in populations between the two studies (military participants in the PTSD dataset, general population in the arachnophobia dataset).
\end{enumerate}  
The results of the PTSD prediction demonstrate strong performance in PCL-M prediction. In the binary classification setting, the model reached an accuracy of 86\%, substantially surpassing the baseline model (52\% accuracy). Importantly, the confusion matrix (Fig. \ref{fig:confusion_matrix}) indicates no systematic bias toward either class, supporting the model's robustness. For continuous prediction of PTSD severity, the model achieved a mean absolute error (MAE) of 5.6 which represents 50\% of the baseline error of 11.2. Accounting for sex as a covariate, the model also outperforms a sex-specific baseline by 48\%, further supporting the utility of sex-aware modeling.

\begin{figure*}[t]
        \centering
    \includegraphics[width=0.5\linewidth]{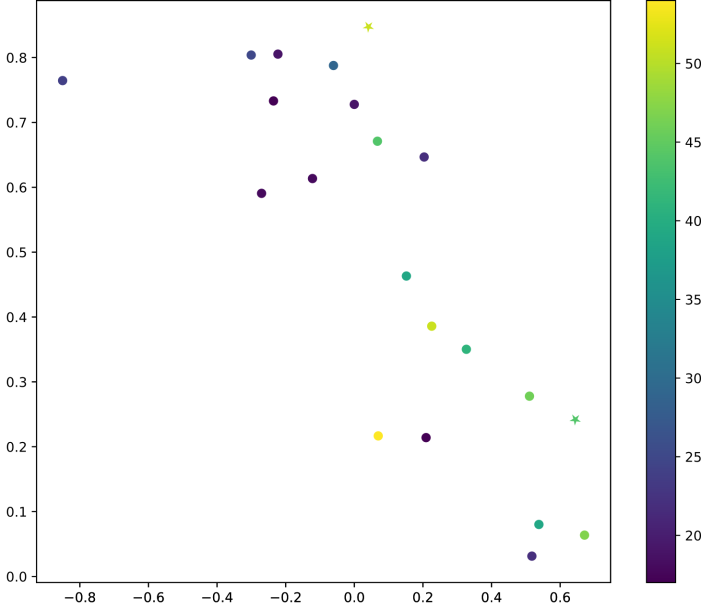}
    \caption{Two-dimensional representation of our model's inputs. The 2 quantitative indicators used in this study are represented as follows: the \textit{fear response slope} is on the x-axis and the \textit{initial fear response} is on the y-axis; the Boolean variable \textit{sex} is shown via the marker shape (a circle for a male and a star for a female). The circled subjects correspond to the two subjects analyzed in the Discussion.}
    \label{fig:twoDplot}
\end{figure*}

The model leverages a compact feature set comprising two continuous predictors (initial fear response and fear response slope) and sex which enables two-dimensional visualization (Fig. \ref{fig:twoDplot}). These features emerge as significant quantitative indicators of PTSD severity. Specifically, subjects with a high initial fear response and a flat or negative fear response slope tend to exhibit lower PCL-M scores, suggesting lower symptom severity. In contrast, those with low initial responses and steeply increasing fear responses tend to present with higher PCL-M scores. These trends are consistent with previous findings on anxiety sensitivity and emotional reactivity in PTSD populations \cite{Bar15}, and aligns with physiological models of PTSD that highlight dysregulation in both fear and cortisol responses \cite{Mae17}.
Two outlier cases merit attention (Fig. \ref{fig:twoDplot}). The first, a "bridge subject" (male, PCL-M = 39), lies between two clusters in latent space. His predicted probability distribution over PCL-M shows two local maxima (Fig. \ref{fig:predictions-proba}), indicating model uncertainty. The second, the countercyclical subject (female, PCL-M = 46), diverges from the cohort pattern. Despite a high initial fear response and flat slope, this participant reports high symptom severity. This could reflect data issues, sex-based variability, or comorbidities since she is diagnosed with MDD. As depression can elevate PCL-M scores \cite{coo05, ger07}, this highlights the need for careful differential diagnosis and possibly sex-specific models, although the small number of female participants (see Table \ref{tab:demographics}) limits such analysis here.

Taken together, these results support the potential of static fear response features as informative, low-dimensional markers for PTSD severity. However, they also highlight the need for more comprehensive clinical validation. Specifically, high PCL-M scores,particularly when not accompanied by formal PTSD diagnoses should be followed by clinical interviews using validated instruments such as the Clinician-Administered PTSD Scale for DSM-5 (CAPS-5), which is more specific to PTSD and less prone to confounding by depression \cite{resick2023}. Ultimately, future work should aim to validate this modeling approach using CAPS-5 as a ground truth and explore its utility in automated differential diagnosis across PTSD and related affective disorders.

The main limitation of this work lies in the small number of subjects in the PTSD cohort. In addition, PTSD prevalence and trauma exposure patterns vary widely across regions and cultures, as noted in the Introduction. This raises important questions about the cross-cultural generalizability of psychophysiological biomarkers and machine learning–based detection systems. While core autonomic responses to trauma (such as elevated heart rate and electrodermal activity) have a biological basis, their manifestation and interpretation can be influenced by cultural, environmental, and contextual factors.
Our sample was also heavily skewed toward male participants, with 90.5\% identifying as male. While research has documented sex differences in fear expression and processing (e.g., \cite{Bau23}), evidence for sex-specific differences in physiological responses to fear over the signals studied in this paper remains mixed. Furthermore, sex differences may be mediated by other variables, including the type and context of trauma, which are themselves unevenly distributed across populations. These factors may limit the generalizability of our findings to female participants.
Understanding how sex, culture, and trauma type interact with physiological responses is crucial for developing machine learning models that are robust and equitable across populations. Future research should examine whether physiological signatures of PTSD are stable across different demographic and cultural groups, or whether specific calibration such as domain adaptation or transfer learning is required to ensure broader applicability.
Finally, a specific immersive simulation related to trauma appears to be needed for PTSD severity estimation using this technique, which is a cumbersome procedure. In the same line, the strong performance of our PTSD prediction model, especially when leveraging transfer learning from a phobia dataset, may in part be attributable to similarities between the fear reaction of phobic subjects and the physiological response profiles to combat-related events in the immersive scenario, which likely induced acute fear and threat-related responses. These responses could be characterized by similar increases in heart rate and electrodermal activity. This partial alignment in autonomic response may have facilitated the success of the model transfer. It remains an open question whether such transfer would be equally effective in cases of PTSD resulting from other forms of trauma.

\section*{Conclusion}

This study hints at the possibility of using transfer learning from a specific phobia physiological dataset to accurately detect PTSD status and estimate PTSD severity. This approach may serve as a foundation for future tools intended for clinical use, pending further validation in larger and more diverse sample considering our cohort size. By leveraging a compact and interpretable feature set, the proposed model achieves high performance in both classification and regression tasks. Additionally, the model’s ability to express uncertainty improve its clinical relevance. These findings support the use of wearable-compatible physiological data for PTSD screening and followup.

\section*{Acknowledgements}
We thank the Army Research Laboratory for trusting us with these data and for the time they dedicated to answering our questions about the protocol.

\section*{Author contribution}
All authors approved the final version of the manuscript. 

\noindent GP and NR performed the data curation, machine learning design, experiments and analysis.

\noindent GP and NR wrote the manuscript with input from all the authors.

\noindent DF, HR, TR and LE did not perform data processing or algorithm validation.

\noindent HR, TR and LE provided the data extraction and preprocessing scripts used in this work.

\noindent HR is the principal investigator of the study used in this paper.

\section*{Data availability}
The data cannot be shared publicly because of US Army Research Laboratory Human Research Protection Program IRB restrictions. The data for this study are stored in a secured laboratory facility in physical form, as well as in an online secured data repository. Data sharing agreements may be put in place for researchers who meet the criteria for access to confidential data. Interested parties may contact Bianca Dalangin (\texttt{bdalangin@dcscorp.com}).

\section*{Funding}

The research study used in this paper was sponsored and funded by the Department of Defense, U.S. Combat Capabilities Development Command’s Army Research Laboratory, Human Research and Engineering Directorate.\\

\section*{Competing interests}
\noindent DF is the CEO and owner of MyndBlue. NR and GP are employees of MyndBlue.

\noindent HR, TR and LE declare no competing interests.

\newpage

\phantomsection
\section*{Supplementary material} 
\label{Supplementary_material}

\subsection*{Tables}

\begin{table}[h!]
\begin{center}
\begin{tabular}{|l|l|}
\hline

GAMESTART & FLASHBANG  \\\hline\rowcolor{lightgray}
AIRRAIDSIREN & EXPLOSION \\\hline
M41 & MP5  \\\hline\rowcolor{lightgray}
MRPEANUTBUTTER Activated & MRPEANUTBUTTER deactivated \\\hline

ABRAMS Activated & \\\rowcolor{lightgray}
ABRAMS Entered INNERRADIUS & ABRAMS Left INNERRADIUS \\
ABRAMS Entered MAXIMUMATTENUATION & ABRAMS Left MAXIMUMATTENUATION  \\\hline\rowcolor{lightgray}

COPCOPTER Activated & COPCOPTER Fired \\
COPCOPTER Entered MAXIMUMATTENUATION & COPCOPTER Left MAXIMUMATTENUATION \\\rowcolor{lightgray}
COPCOPTER Entered INNERRADIUS & COPCOPTER Left INNERRADIUS \\\hline

DUMMYTANK Activated & \\\rowcolor{lightgray}
DUMMYTANK Entered MAXIMUMATTENUATION & DUMMYTANK Left MAXIMUMATTENUATION \\
DUMMYTANK Entered INNERRADIUS & DUMMYTANK Left INNERRADIUS \\\hline\rowcolor{lightgray}

PATRIOT Activated & \\
PATRIOT Entered MAXIMUMATTENUATION & PATRIOT Left MAXIMUMATTENUATION \\\rowcolor{lightgray}
PATRIOT Entered INNERRADIUS & PATRIOT Left INNERRADIUS \\\hline

PUMA Activated & \\\rowcolor{lightgray}
PUMA Entered MAXIMUMATTENUATION & PUMA Left MAXIMUMATTENUATION \\
PUMA Entered INNERRADIUS  & PUMA Left INNERRADIUS  \\\hline\rowcolor{lightgray}

\end{tabular}
\caption{List of events occurring during the `High Stress Army' virtual immersive simulation. There are different types of events, including visual and sound stimuli. All events are related to a military context, except for `GAMESTART' (beginning of the simulation) and `MRPEANUTBUTTER' (a dog barking). For most of the events, 3 types of statuses were possible during the simulation.\\
- `Activated': The participant has triggered the event and the object (e.g. helicopter or tank) has started moving\\
- `Entered MAXIMUMATTENUATION': The participant is close enough where  sound level is at maximum.\\
- `Left MAXIMUMATTENUATION': The participant is no longer close enough to hear the object at a maximum sound level.\\
- `Entered INNERRADIUS': Due to movement of the object and/or the participant, the participant is approaching the object (or has entered a specified radius around the object).\\
- `Left INNERRADIUS': Due to movement of the object and/or the participant, the participant is moving farther away from the object (or has left a specified radius).
}
\label{tab:event}
\end{center}
\end{table}

\begin{table}[!h]
\begin{center}
\bgroup
\def\arraystretch{1.75}
\begin{tabular}{|l|l|}
\hline
\rowcolor{white}  \textbf{Name}   &  \textbf{Formula} \\
  \hline
  \hline
\rowcolor{lightgray}
  Nmean   & $Nmean_t = \frac{1}{20\cdot fs}\sum_{t \geq t_0}^{t <t_1} \tilde{\phi}_t$ \\
  \hline
  Nstd & $\frac{1}{20\cdot fs} \sum_{t \geq t_0}^{t <t_1} (\tilde{\phi}_t - Nmean_t)^2$ \\
  \hline
\rowcolor{lightgray}
  \hline
  Ndiff1 & $\frac{1}{20\cdot fs}\sum_{t \geq t_0}^{t <t_1-1} |\tilde{\phi}_{t+1} - \tilde{\phi}_t|$ \\ 
  \hline
  Ndiff2 & $\frac{1}{20\cdot fs}\sum_{t \geq t_0}^{t <t_1-2} |\tilde{\phi}_{t+2} - \tilde{\phi}_t|$ \\ \hline
\end{tabular}
\egroup
\end{center}
    \caption{Formulas describing the channelwise computation of 4 features related to the fear response in 20-seconds windows of physiological data. These features were originally used for emotion recognition \cite{Pic01}. Here $\tilde{\phi}_{t_0}, \ldots, \tilde{\phi}_{t_1}$, for $t_1- t_0 = 20$ seconds, is a subjectwise normalized physiology window corresponding to one of the 3 channels.}
    \label{tab:feature_construction}

\end{table}

\begin{table}
    \centering
    \begin{tabular}{|l|l|}
        \hline
        \textbf{parameter} &  \textbf{value} \\\hline\hline
        \rowcolor{lightgray}
        n\_units & 16 \\
        \hline
        depth & 6 \\
        \rowcolor{lightgray}
        \hline
        hidden activations & rectified linear unit (ReLU) \\
        \hline
        final activation & sigmoid \\
        \rowcolor{lightgray}
        \hline
        epochs & 100 \\
        \hline
        batch\_size & 512 \\
        \rowcolor{lightgray}
        \hline
        learning\_rate & 0.01 \\
        \hline
        optimizer & stochastic gradient descent with momentum \\
        \rowcolor{lightgray}
        \hline
        momentum coefficient & 0.9 \\
        \hline
        weight\_decay & 0.001 \\\hline
    \end{tabular}
    \caption{Phobia-like stress model parameters and training hyperparameters. This model is an MLP, so its architecture is described in terms of width, depth, and activations. The learning scheme is described in terms of the optimizer and learning hyperparameters.}
    \label{tab:hyperparameters}

\end{table}

\begin{table}[h!]
\begin{center}
\bgroup
\def\arraystretch{1.75}
\begin{tabular}{|l|l|l|l|}
\hline
\rowcolor{white}
\textbf{Kernel name} & $\mathbf{k(x, y)}$  &\textbf{ type of feature} & \textbf{parameters} \\
\hline \hline \rowcolor{lightgray}
   Radial Basis Function ($RBF$) & $\frac{1}{\sqrt{2\pi\nu}} e^{((x-y)/\nu)^2}$ & continuous  & bandwidth $\nu$ \\ \hline
    Bernoulli Kernel ($BK$)described & $(1-x)(1-p) + xp$ & binary & probability $p$\\  \hline
\end{tabular}
\egroup
\end{center}
\caption{The building-block kernels used in this paper. For simplicity, we choose the $RBF$ kernel for continuous variables and the $BK$ kernel for binary variables.} 
\label{tab:kernels}
\end{table}

\subsection*{MKDE implementation}

\subsubsection*{Model construction} 
The MKDE model learns the probability distribution on the possible outcomes. It is a density estimation model, where the density functions are estimated over each variable individually and then aggregated; these can be generalized probability distributions (e.g., Gaussian, Bernoulli, Polynomial) used in a uniform fashion. 
For this model, we choose a Gaussian kernel for the 2 continuous variables, namely, the \textit{fear response slope} and \textit{initial fear response}, and a Bernoulli kernel for the binary variable \textit{sex}. Finally, the distribution of the PCL-M label is estimated via a Gaussian kernel. The model outputs a probability distribution on the possible values for the PCL-M score, i.e., between 17 and 85. The details of the construction of MKDE are provided below.\\

Since the dataset contains only 21 subjects, separation into training and test sets is performed via a leave-one-out (LOO) validation scheme. Thus, for each subject, a new model is trained with the data of this subject as the test set, and the data of all other subjects as the training set. The raw output of the MKDE models is a probability distribution over all possible values for PCL-M; in other words, the curve $P(y)$ is defined for $y \in [17, 85].$ This raw prediction can be reduced to a single PCL-M prediction as follows: $y_{pred} = \text{argmax}_{y\in [17,85]} \ P(y).$ Finally, we can further reduce this PCL-M prediction by applying a threshold of 36 to distinguish between PTSD patients and non-PTSD patients, as discussed in the PTSD dataset subsection:
$$PTSD_{pred} = \begin{cases}
1 &\text{if $y_{pred} \geq 36$,}\\
0 &\text{if $y_{pred} < 36$.}
\end{cases}$$

\subsubsection*{Overview} Our MKDE model is a general ML model suited for small datasets that closely follows the general form of the classical kernel density estimation (KDE). This model computes the kernel regression function described in \cite{hyn96, nad64, wat64} in its general form. Let $(X, y)$ be a small dataset to be used for training, where $X$ is an array of $M$ samples of dimension $d$ and $y$ is a vector containing the $M$ labels associated with the samples in $X$, and let $(X_{test}, y_{test})$ be a testing set of the same dimension with $M_{test}$ samples.

In summary, the principles of the MKDE model are as follows. First, a kernel of dimension $d+1$ is chosen, for instance, a product of a Gaussian kernel if all the variables are continuous. Then, an MKDE model is fit to estimate the probability distribution of the $d+1$-dimensional dataset $\hat{X} = [X, y]$, obtained by adding $y$ as an extra feature to $X$. This results in a probability distribution
$$ P: \mathbb{R}^{d+1} \rightarrow \mathbb{R}.$$
Finally, the output of this model on a sample $x$ is a probability distribution on $\mathbb{R}$, defined as
$$ MKDE(x)(y) = P([x, y]).$$
In the case of missing variables, an estimation of the probability distribution over $y$ can still be obtained through marginalization: if the test sample is $x = [x_{partial}, x_{missing}]$, where $x_{missing}$ contains $d - d_{partial}$  placeholder values, then
\begin{equation}
MKDE(x)(y) = \int_{\mathbb{R}^{d - d_{partial}}} P([x_{partial},\tilde{x}, y]) d\tilde{x}. \label{eq:margin}
\end{equation}
This is essentially a reformulation of \cite[eq. (4.1)]{hyn96} in our setting.
On a lower level, the MKDE does not estimate the probability distribution itself, since this would imply choosing a grid over $\mathbb{R}^{d+1}$, which would lead to the accumulation of errors and a low computational efficiency. The estimation procedure is described below.

\subsubsection*{Preprocessing and kernels} In this work, we consider only 2 types of stationary kernels, i.e., radial basis function (RBF) and Bernouilli kernel (see the formulas in Table \ref{tab:feature_construction} in the Supplementary Material). To reduce the number of hyperparameters of our model to a single number $\sigma$, representing the bandwidth of the $d+1$-dimensional kernel, we assume that the continuous variables are standardized. Then, given a dataset with $d_{\mathbb{R}}$ continuous variables and $d_{\mathcal{B}}$ binary variables such that $d = d_{\mathbb{R}} + d_{\mathcal{B}}$, the standard kernel with parameter $0<\sigma<0.5$, denoted by $K(\sigma)$, is defined as the product kernel 
$$ K(\sigma) = \prod_{i=0}^{d_{\mathbb{R}}-1} RBF(\sigma) \times\prod_{i=0}^{d_{\mathcal{B}}-1} BK(0.5 + \sqrt{0.25 - \sigma^2}).$$
The standard kernel is also simply denoted as $K$ if there is no ambiguity in the value of $\sigma$. Without loss of generality, we assume that the features in the dataset are ordered with continuous variables before binary variables, which is always true up to a reordering of the variables.

\subsubsection*{Training and prediction} Given $X$ and $y$ as above, let $K_X$ and $K_y$ be the components of $K(\sigma)$ corresponding to $X$ and $y$, respectively. The MKDE model's prediction function with parameter $\sigma$ is defined as follows:
$$ P(y | X = x) =\frac{1}{M} \sum_{m=0}^{M-1} K_y(y, y_m) * K_X(x, x_m),$$

which is an approximation of the marginalization formula \eqref{eq:margin} for finite samples instead of distributions. The parameter $\sigma$ is chosen as 
$$ \sigma = \text{argmax} \prod_{m=0}^{M-1} P(y_m | X=x_m).$$ 

In practice, the best fit for $\sigma$ over the training set is determined through a grid search for $\sigma \in (0, 0.5)$, since this procedure is inexpensive considering the size of our dataset.

\end{document}